%% file: Bib03.PANSER_Generation.tex
\documentclass{article}
\usepackage{cite}
\usepackage{graphicx}
\usepackage{hyperref}
\usepackage{textcomp}
\usepackage{acro}
\usepackage{tabularx} 
\usepackage{diagbox} 
\usepackage{multirow}
\usepackage{amssymb}
\usepackage{amsmath}
\usepackage{nicefrac}
\usepackage{mathrsfs}
\usepackage{booktabs}
\usepackage[table]{xcolor}
\usepackage{comment}

\def\netname{MerGen}

\newcommand{\etal}{{\em et al.\ }}

\input{acronyms}

\begin{document}

\title{\netname: Micro-electrode recording synthesis using a generative data-driven approach}

\author{Thibault Martin$^1$ \href{https://orcid.org/0000-0001-5655-0923}{\includegraphics[scale=0.65]{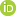}} 
\and Paul Sauleau$^{1,2}$
\and Claire Haegelen$^{1,3}$ 
\and Pierre Jannin$^1$ \href{https://orcid.org/0000-0002-7415-071X}{\includegraphics[scale=0.65]{src/logos_PP/orcid.png}}, \and John~S.H.~Baxter$^1$ \thanks{Corresponding author email: john.baxter@univ-rennes.fr} \href{http://orcid.org/0000-0003-3548-4343}{\includegraphics[scale=0.65]{src/logos_PP/orcid.png}}
}

\date{
$^1$~Laboratoire Traitement du Signal et de l'Image (INSERM UMR 1099), University of Rennes (Rennes, France)\newline
$^2$~Rennes University Hospital Centre (Rennes, France)\newline
$^3$~Lyon University Hospital Centre (Lyon, France)~}

\markboth{\hskip25pc}
{Martin \MakeLowercase{\textit{et al.}}: \netname: Micro-electrode recording synthesis using a generative data-driven approach}

\maketitle

\input{01_Abstract}
\input{02_Introduction}

\input{03_Network}
\input{04_M_M}

\input{05_Results}
\input{06_Discussion}
\input{07_Conclusion}

\bibliographystyle{IEEEtran}
\bibliography{Bib03.PANSER_Generation} 

\end{document}

%% file: acronyms.tex
\DeclareAcronym{DL}{
	short = DL,
	long  = Deep Learning,
	tag = abbrev
}

\DeclareAcronym{MRI}{
	short = MRI,
	long  = Magnetic Resonance Imaging,
	tag = abbrev
}

\DeclareAcronym{STN}{
	short = STN,
	long  = Subthalamic Nucleus,
	tag = abbrev
}

\DeclareAcronym{STFT}{
	short = STFT,
	long  = Short Term Fourier Transform,
	tag = abbrev
}

\DeclareAcronym{MER}{
	short = MER,
	long  = Micro-electrode Recording,
	tag = abbrev
}
\DeclareAcronym{EDT}{
    short = EDT,
    long = Estimated Distance from Target,
    tag = abbrev
}
\DeclareAcronym{HAC}{
    short = HAC,
    long = Hierachical Agglomerative Clustering,
    tag = abbrev
}

\DeclareAcronym{DBS}{
	short = DBS,
	long  = Deep Brain Stimulation,
	tag = abbrev
}
\DeclareAcronym{MUA}{
	short = MUA,
	long  = Multi-Unit Activity,
	tag = abbrev
}
\DeclareAcronym{BUA}{
	short = BUA,
	long  = Background Unit Activity,
	tag = abbrev
}
\DeclareAcronym{PD}{
	short = PD,
	long  = Parkinson's Disease,
	tag = abbrev
}

\DeclareAcronym{VQVAE}{
	short = VQ-VAE,
	long  = Vector Quantized Variational Auto-Encoder,
	tag = abbrev
}

\DeclareAcronym{GAN}{
	short = GAN,
	long  = Generative Adversarial Network,
	tag = abbrev
}

\DeclareAcronym{TSTR}{
	short = TSTR,
	long  = Train-on-Synthetic-only Test-on-Real,
	tag = abbrev
}

\DeclareAcronym{UMAP}{
    short = UMAP,
    long =  Uniform Manifold Approximation and Projection for Dimension Reduction,
    tag = abbrev
}

\DeclareAcronym{TRTR}{
	short = TRTR,
	long  = Train-on-Real-only Test-on-Real,
	tag = abbrev
}

\DeclareAcronym{TRSTR}{
	short = TRSTR,
	long  = Train-on-Real-and-Synthetic Test-on-Real,
	tag = abbrev
}
\DeclareAcronym{BACC}{
	short = BACC,
	long  = Balanced Accuracy,
	tag = abbrev
}

\DeclareAcronym{SVS}{
    short = SVS,
    long = Singing Voice Synthesis,
    tag = abbrev
}

\DeclareAcronym{TTS}{
    short = TTS,
    long = Text To Speech,
    tag = abbrev
}

\DeclareAcronym{EMA}{
    short = EMA,
    long = Exponential Moving Average,
    tag = abbrev
}

\DeclareAcronym{EEG}{
    short = EEG,
    long = Electroencephalography,
    tag = abbrev
}
\DeclareAcronym{EMG}{
    short = EMG,
    long = Electromyography,
    tag = abbrev
}
\DeclareAcronym{ECG}{
    short = ECG,
    long = Electrocardiography,
    tag = abbrev
}

%% file: 01_Abstract.tex
\begin{abstract}

The analysis of electrophysiological data is crucial for certain surgical procedures such as deep brain stimulation, which has been adopted for the treatment of a variety of neurological disorders. During the procedure, auditory analysis of these signals helps the clinical team to infer the neuroanatomical location of the stimulation electrode and thus optimize clinical outcomes. This task is complex, and requires an expert who in turn requires significant training.
In this paper, we propose a generative neural network, called MerGen, capable of simulating \textit{de novo} electrophysiological recordings, with a view to providing a realistic learning tool for clinicians trainees for identifying these signals.
We demonstrate that the generated signals are perceptually indistinguishable from real signals by experts in the field, and that it is even possible to condition the generation efficiently to provide a didactic simulator adapted to a particular surgical scenario. The efficacy of this conditioning is demonstrated, comparing it to intra-observer and inter-observer variability amongst experts. We also demonstrate the use of this network for data augmentation for automatic signal classification which can play a role in decision-making support in the operating theatre.

\vspace{1mm}

\end{abstract}

%% file: 02_Introduction.tex
\section{Introduction}

The goal of generative modeling is to capture the salient aspects of a data distribution and reproduce them in a set of new instances indistinguishable from real data. This task has recently become widely democratized since the emergence of Generative Adversarial Networks (GANs). In medicine, this approach has already demonstrated utility for generating imaging data for a wide range of applications  \cite{yi_generative_2019}.

Moving from images to audio data, the landscape changes considerably. While GANs have facilitated significant advancements in image data, characterized primarily by spatial relationships, the generation of time-series data introduces unique challenges due to the need to capture long-range dependencies. In the field of general audio task generation, substantial progress has been made in autoregressive architectures, especially self-attention mechanisms \cite{vaswani_attention_2017}. These developments have revolutionized sequential data modelling while overcoming the limitations of previous methods such as recurrent neural networks, making it possible to consider the high fidelity synthesis of long temporal series. This is especially true in text-to-speech, singing voice synthesis, or music composition \cite{shi_survey_2021}.

For biomedical signals, the emergence of data synthesis aligns well with the increasing demand for clinical data necessary for increasingly complex data-driven algorithms and for providing educational support for training novice clinicians. The generation of biomedical time series is a relatively new field and recent efforts have been concentrated on generating short-length series in specific modalities such as electrocardiography \cite{delaney_synthesis_2019}, electroencephalography \cite{hazra_synsiggan_2020} or electromyography \cite{golany_pgans_2019}. The data generated in these studies suffer from the technical limitations, i.e. the ability to generate short segments of data, averaging a few hundred samples, which offers a poor simulation from a didactic support perspective. In addition, these studies are limited by the lightness of their protocol for qualitative evaluation of the generated data. These works demonstrate the potential for data-driven models to simulate realistic events while preserving patient anonymity. This can help with data accessibility, which can be hampered by privacy concerns. 

\subsection{Micro-electrode recordings for deep brain stimulation electrode implantation}

\ac{DBS} is an interventional treatment for various neurological and neurodegenerative disorders, notable Parkinson's disease \cite{harmsen2022we}. This treatment involves the positioning of an electrode at a particular region of the patient's subcortical anatomy. This region needs to be very precisely targeted which means that uncertainties caused by limited pre-operative image resolution as well as small errors in implementing the pre-operative plan need to be accounted for using interoperative data acquisition.

Promising results have been observed for intraoperative imaging such as interventional MRI. However, due to its low cost and high accessibility, as well as added value in post-operative results, \ac{MER} analysis is very popular for guiding \ac{DBS} electrode implantation \cite{ maiti_intra-operative_2016}.
By capturing variations in the local extracellular electric field, \ac{MER} provides information about the tissue surrounding the probe. These signals share some characteristics with other electrophysiological signals, as they are information-dense and capture quasi-random physiological events which can be difficult to distinguish from noise or artifacts.

But despite their advantages, the analysis of these signals is relatively complex, requiring the assistance of an expert neurophysiologist for several hours during electrode implantation. Furthermore, the literature shows that an extended analysis time (and thus a longer intervention) is associated with an increased risk of hemorrhage \cite{gorgulho_incidence_2005}.
These constraints greatly motivate the development of a method capable of simulating this analysis stage of the surgical workflow, in order to democratize the understanding of these data, provide training platforms for novice neurophysiologists to learn the basics of these skills outside of the operating theatre, and thus facilitate the use of \ac{MER} in the clinical context. Given the subjective nature of auditory analysis, it is particular important that generated signals be both realistic and highly variable, exposing novices to a wider array of situations without them becoming subconsciously habituated to a particular signal.

\subsection*{Contributions}
This paper presents MerGen, a generative designed for real-time simulation of \ac{MER} signals. This network aims to provide a learning resource for trainees in the use of said signals for identifying neuroanatomy during \ac{DBS} electrode implantation. The contribution of this paper also include a double qualitative evaluation. First, the perceived realism of the simulated signals is assessed. Second, MerGen's data augmentation capacities is evaluated in the context of \ac{MER} signal classification, showing that its realism is more than perceptual.

%% file: 03_Network.tex
\begin{figure}[t]
\centering
    \includegraphics[width=\textwidth,keepaspectratio]{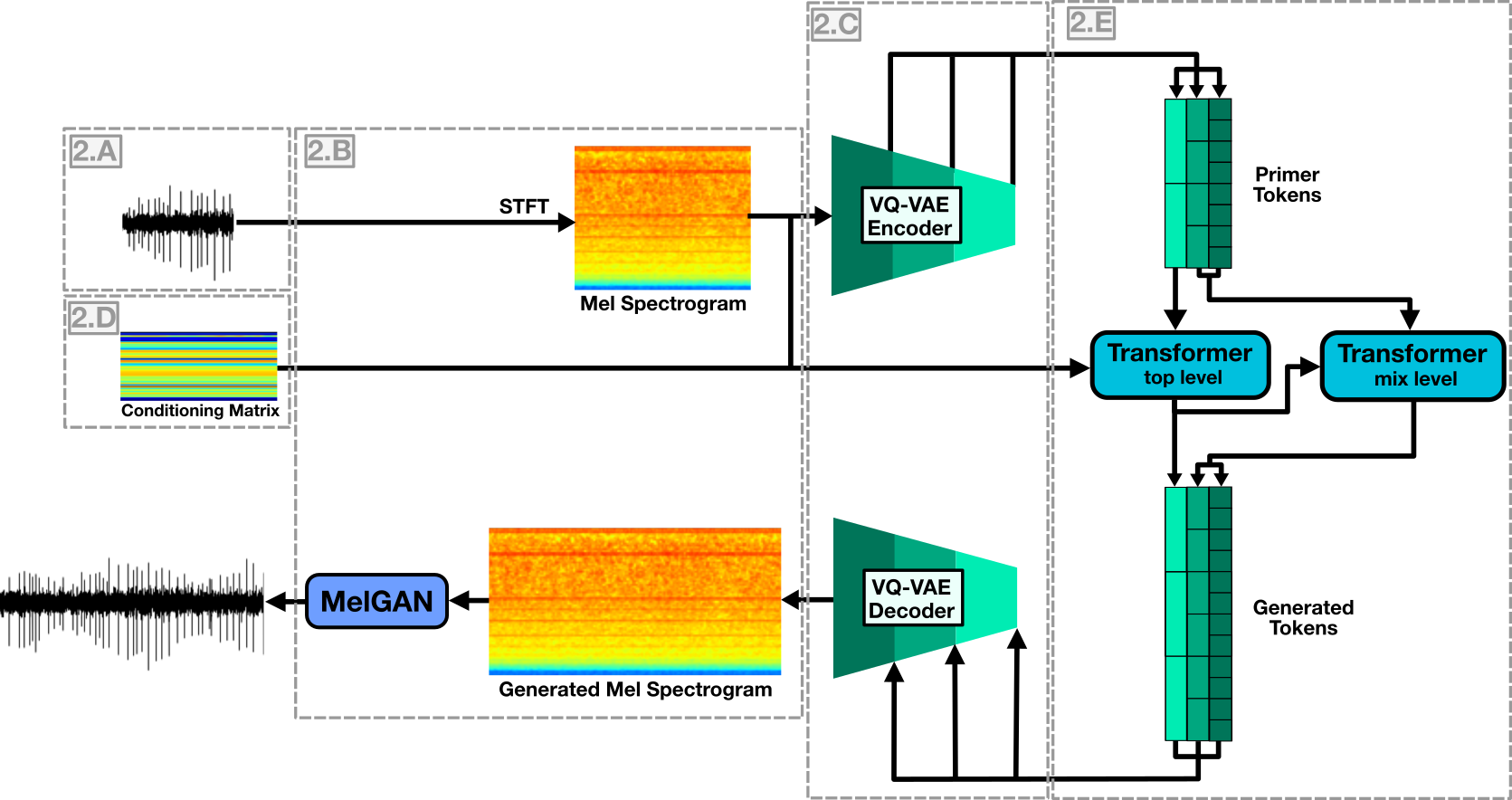}
    \caption{MerGen: Architecture with corresponding method sections.}
\label{fig:mergenschema}
\end{figure}

\section{Material and Methods}

MerGen's architecture is shown in Figure \ref{fig:mergenschema} and consists of three separately-learnt components:
\begin{enumerate}
    \item A GAN-based network for inverting spectrograms used to represent the signal (Section \ref{sec:melgan};
    \item A \ac{VQVAE} network for representing the signal as a sequence of discrete tokens (Section \ref{sec:vq-vae}); and
    \item A cascaded transformer model for generating new token sequences (Section \ref{transformer_section}).
\end{enumerate}

\subsection{Signal acquisition and pre-processing}

The \ac{MER} data used in this study was collected from a cohort of 63 patients undergoing \ac{DBS} surgery for \ac{PD} at the Rennes University Hospital Centre. 
Each patient was implanted with either one or two electrodes for single or bilateral \ac{STN} stimulation. Data recording was performed using a Ben-Gun configuration, with a simultaneous progression of 2 to 3 recording micro-electrodes. The stimulation target coordinates were preoperatively estimated and 20-40 \ac{MER}s collected between 7 mm before and 2 mm after the target.

Data acquisition was performed using the KeyPoint.NET software on a Dantec Keypoint G recording station. The data was sampled using the same parameters used by the clinical team for intraoperative auditory analysis. Recording was carried out at a sampling frequency of 24kHz, amplified (gain 10,000). A notch filter was applied at 50Hz, along with 500-5000Hz high-pass and low-pass filters.
The data acquired for this study are covered by the ethical authorization of declaration n2205295 v0, in accordance with CNIL reference methodology MR004 and approved by the ethics committee at the Rennes University Hospital Centre.

We have compiled a total of 8,630 3-second signals each annotated with the acquisition depth, \ac{EDT} (the target being the proximal border of the \ac{STN}), and neurophysiologist's intraoperative score. This score reflects the perceived \ac{STN} activity on an ordinal scale (shown in Table \ref{table:annotation_prevalence}) and was determined by an expert neurophysiologist during \ac{DBS} electrode implantation.

\begin{table}[!ht]
\caption{Total lengths of recordings associated with different levels of activity characteristic of the \ac{STN}}
\begin{tabularx}{\linewidth} { 
   >{\raggedright\arraybackslash}X 
  | >{\centering\arraybackslash}X 
  | >{\centering\arraybackslash}X 
  | >{\centering\arraybackslash}X 
  | >{\centering\arraybackslash}X }
\toprule
Annotation & Level of Activity & Amount in Training Set & Amount in Testing Set & Total length \\
 & Characteristic of the STN & ($n=51$ patients) & ($n=12$ patients) &\\
\hline
$-$ & Missing Activity &00h04m33s&00h02m57s& 00h07m30s \\ 
$-+$ & Very low Activity &01h45m30s&00h24m27s& 02h09m57s \\ 
$+$ & Low Activity &01h30m12s&00h20m00s& 01h50m12s \\ 
$++$ & Moderate Activity &01h26m03s&00h17m33s& 01h43m36s\\ 
$+++$ & Strong Activity &01h08m33s&00h11m42s& 01h20m15s \\ 
\bottomrule
\end{tabularx}

\label{table:annotation_prevalence}
\end{table}

The patients were split into a training set and a testing set to avoid data leakage, as signals arising from the same patient are likely to be correlated, especially with regards to artifacts. Once of the patients were split into these sets, they were then decomposed into a large number of signals. The distribution of signals within these two sets is also shown in Table \ref{table:annotation_prevalence}. 
After inspection of the signal amplitude distribution, amplitudes exceeding $[-250mV:+250mV]$ were considered artifactual and were clipped. The signal amplitude was then normalized to lie within $[-1,+1]$.

\subsection{Signal representation and reconstruction}\label{sec:melgan}
In many acoustic signal analysis methods, signals are represented via a spectrogram allowing for a richer representation of the signal to be immediately available to the network. The \ac{MER}s were transformed into a spectrogram by a \ac{STFT} algorithm, with a Hann window size of 1024 samples, a hop length of 256, and the frequency scale mapped to the Mel scale in 80 distinct bands so each 3-second signal produces a matrix of size (80x280) as output. After minimal thresholding at $1\mathrm{e}^{-5}$, the Mel-spectrogram intensity was logarithmic scaled. 

As our generation network takes (and thus produces) data in this spectrogram domain, it is necessary to invert this process to construct a fully time-domain acoustic signal. However, these spectrograms retain only magnitude (and not phase) information necessary to reconstruct the signal.
The standard Griffim-Lim spectrogram inversion algorithm \cite{griffin_signal_1984} showed poor results in our case, we thus opted for the use of a lightweight data-driven model, a MelGAN network \cite{kumar_melgan_2019}, which could be adapted for our particular problem domain. The MelGAN architecture is a feed-forward convolutional network for phase reconstruction, trained in a GAN setup consisting of a generator $(G)$ and three discriminator $(D)$ networks. $G$ is designed to take a Mel-spectrogram as input and output the corresponding audio waveform by applying a series of transposed convolutions followed by dilated convolutions. These operations have a large receptive field and are thus capable of producing time series with a large number of samples. The discriminators ($D1$, $D2$, $D3$) are three identical networks, which take a signal at a particular scale and outputs whether or not the signal is real or was produced by $G$. Each network uses an average-pooled signal with kernels set at 1, 2 and 4 respectively. This discriminator ensures that the network generates realistic signals with the same magnitude components.
The GAN was trained according to the loss functions from in the original MelGAN paper \cite{kumar_melgan_2019}:

\begin{equation*}\label{eq:melgan_loss}
\begin{aligned}
    &\underset{D_k}{\min}\;\mathbb{E}_{x}\left[\min(0,1-D_k(x))\right]  + \mathbb{E}_{s,z}\left[\min(0,1+D_k(G(s,z)))\right] \\
    &\min_{G} \; \mathbb{E}_{s,z}\left[\sum_{k=1,2,3}-D_k(G(s,z))\right] 
\end{aligned}
\end{equation*}
%
Training was carried out over 2000 epochs using an Adam optimizer with a batch size of 16 and learning rate of $1\mathrm{e}^{-4}$.

\subsection{Signal reduction in a discrete latent space}\label{sec:vq-vae}
Modeling the signal's evolution in the spectogram-domain is a complicated task, leading us to use autoregression in a simpler discrete space. To achieve this, spectrograms were compressed into discrete codes using a \ac{VQVAE} architecture similar to Karasinger \cite{liao_karasinger_2021}, reducing each input spectrogram into three levels of representation: \textit{bottom}, \textit{middle}, and \textit{top}. At each level, the data is temporally compressed, by a factor of 1, 2, and 4 respectively. Finally, their output is expressed in the $[-1,1]$ range via a hyperbolic tangent function.

For each level, the encoder outputs are then decomposed into a series of vectors, which together form the set $\pmb{v_l} \in R^{D}$, with $l = 1, ..., L$, where $D$ corresponding to the size of the vector, 80 in our case, and $L$ corresponding to the number of vectors per output, 280, 140 and 70 respectively for each level of decomposition. Each vector describes the signal's abstract content at time $t$.

Quantization of the encoder output is achieved by replacing each $\pmb{v_l}$ vector with its nearest neighbor in an embedding space. This codebook dictionary is designated as $\pmb{e_w} \in R^{D}$, with $w = 1, ..., W$, where $W$ standing for the size of the dictionary, 1024 in our case. Each latent level has its own codebook, optimized during training by the \ac{EMA} method. The quantization operation is described in the following equation, where latent vectors $\pmb{v_l}$ are transformed into quantized vectors $\pmb{v'_l}$:
    $$\pmb{v'_l} = \pmb{e}_{\underset{w}{\arg\min}||\pmb{v}_{l} - \pmb{e_w}||_{2}^{2}}$$
Backpropagation through this quantization was performed using Oord \etal's \cite{oord_neural_2018} \textit{stop-gradient} method. In order to facilitate convergence and avoid a collapse in codebook usage towards a small number of vectors, we used the codebook trick method \cite{dhariwal_jukebox_2020}, where we reset dictionary words if they fall below a minimum frequency ($1e^{-6}$) and replace them with random data from the input batch. Once the data has been discretized, it is passed to a decoder based on that of Liao \etal \cite{liao_karasinger_2021} to reconstruct the input spectrogram.

The loss function used to train the \ac{VQVAE} network was:
\begin{equation*}
\mathcal{L} = \mathcal{L}_{mel} + \mathcal{L}_{wav} + \lambda \mathcal{L}_{commit} + \beta \mathcal{L}_{cdnt}
\end{equation*}
where $\beta$ = 1 and $\lambda$ = 0.25. The losses used are:
\begin{itemize}
\item $\mathcal{L}_{mel}$ : Pixel-by-pixel distance between auto-encoder input and output spectrograms.
\item $\mathcal{L}_{wav}$ : Amplitude distance between the signal reconstruction from the input spectrogram and the autoencoder output spectrogram.
\item $\mathcal{L}_{commit}$ : Distance between encoder output data and mapping after quantization.
\item $\mathcal{L}_{cdnt}$ : Distance between the top latent space and conditioning matrix described in Section \ref{cdtn_features_section}.
\end{itemize}
The network was trained using an Adam optimizer, a learning rate of $5e^{-5}$ for 2000 epochs, and a batch size of 16.

Once the VQVAE was trained, MER signals are decomposed into a sequence of vectors in this 3-level space, each level describing the input signal at different scales. The \textit{bottom} and \textit{middle} levels use a very local scale with a resolution closer to that of the spectrogram. On the other hand, the \textit{top level} describes the high-level features from the input signal. In order to improve auto-encoder training and modulate the nature of the signal generated by an autoregressive model, we have developed a solution to reduce the nature of each \ac{MER} to a conditioning matrix in the same dimensions as the data at the \textit{top level}. As a result, the cost function $\mathcal{L}_{cdnt}$ allows the integration of conditioning as early as the signal vector decomposition stage, providing better results and forcing the latent space to organise the data by physiological context.

\subsection{Conditioning features}\label{cdtn_features_section}

The precise characteristics of \ac{MER} signals heavily depend on context, such as the position of the electrode with respect to the underlying anatomy, the precise set-up of the signal acquisition system, etc... In order to ensure contextually-appropriate tokens are generated, some additional information is required.
The \ac{EDT} and the neurophysiologist's annotation are crucial modalities for sampling from a generator with the aim of simulating the interventional process. However, these parameters are not fully informative. \ac{EDT} annotations for example only indicate the probability of encountering a type of electrical activity rather than fully specifying it. Expert annotations are more explicit in this regard but are subjective. Neither captures notions of artefacts which are often present in the real \ac{MER} signals and should also be present in simulated ones. We therefore considered increasing the amount of conditioning information to facilitate model learning, and improve the consistency and diversity of generated data.
Given the non-stationary nature of the signal, we divided the conditioning information into two categories: \textbf{global conditioning} which is invariant over the entire signal, and \textbf{local conditioning} which describes a short time window. 

The global conditioning includes: 1) the \ac{EDT}, 2) a one-hot vector reflecting the activity annotation, and 3) an artifact profile. 
During certain procedures, electromagnetic interference artifacts have a high salience in the recordings obtained. To prevent the generator from collapsing into the production this single type of noise, we sought to isolate the main classes of noise in an unsupervised manner, and integrate them as conditioning. From a spectrogram representation, this type of artifact is identified by an over-expression of amplitude for certain frequency bands. To isolate these bands, a convolutional horizontal line detection filter was applied to all Mel-Spectrogram representations and the results partitioned by hierarchical agglomerative clustering. Three distinct clusters were found and we included a vector for each in the conditioning matrix.

For local conditioning, the following metrics were calculated over 70 non-overlapping windows within each 3-second recording:
\begin{enumerate}
    \item \textit{Power ratio 100-150Hz / 5-25Hz}: The intensity ratio between these two bands helps to discriminate between the \ac{STN} from the \textit{substantia nigra} often encountered deeper on the electrode trajectory. \cite{wan_review_2018}
    \item \textit{Spike count}: via the WaveClus automatic spike detection tool. \cite{chaure_novel_2018}

    \item \textit{Root Mean Square Amplitude:} $\sqrt{\frac{1}{N} \sum_{i=1}^{N}x_{(i)}^{2}}$
    \item \textit{Curve length:} $\sqrt{\sum_{i=1}^{N-1}\left|x_{(i+1)}-x_{(i)}\right|}$
    \item \textit{Number of zero crossings:} \\
        $\frac{1}{2}\sum_{i=1}^{N-1}\left|\operatorname{sign}(x_{(i+1)}) - \operatorname{sign}(x_{(i)})\right|$
\end{enumerate}
where $x_i$ is the \ac{MER} signal amplitude across length $N$. Each feature was scaled to the $[-1, 1]$ range in the training database.

\subsection{Generating sequences of tokens} \label{transformer_section}

As previously stated, the \ac{VQVAE} model provides a representation of the \ac{MER} signal as a sequence of discrete tokens making it is necessary to model the temporal progression of tokens in these sequences. Autoregressive modeling in those discrete spaces was carried out in a self-supervised approach using a model inspired by that presented by Liao \etal \cite{liao_karasinger_2021} who showed that using two networks to model the 3 latent spaces produced more robust results than using one network per level. We therefore model the prior $p(v')$ according to the following relationship:
\begin{equation*}
p(v') = p(v'_{top}, v'_{mid}, v'_{bottom}) = p(v'_{top})p(v'_{mid},v'_{bottom}|v'_{top})
\end{equation*}
In our case, the prior is modeled by two linear transformer networks organized in a cascade. The first network models the prior $p(v'_{top})$, producing a sequences which will be used as input for the second transformer network, modeling $p(v'_{mid},v'_{bottom}|v'_{top})$. Both networks have been trained using the 3 levels of token series by the \ac{VQVAE} encoder.

\begin{figure}[t!]
\centering  
\includegraphics[width=0.85\linewidth,keepaspectratio]{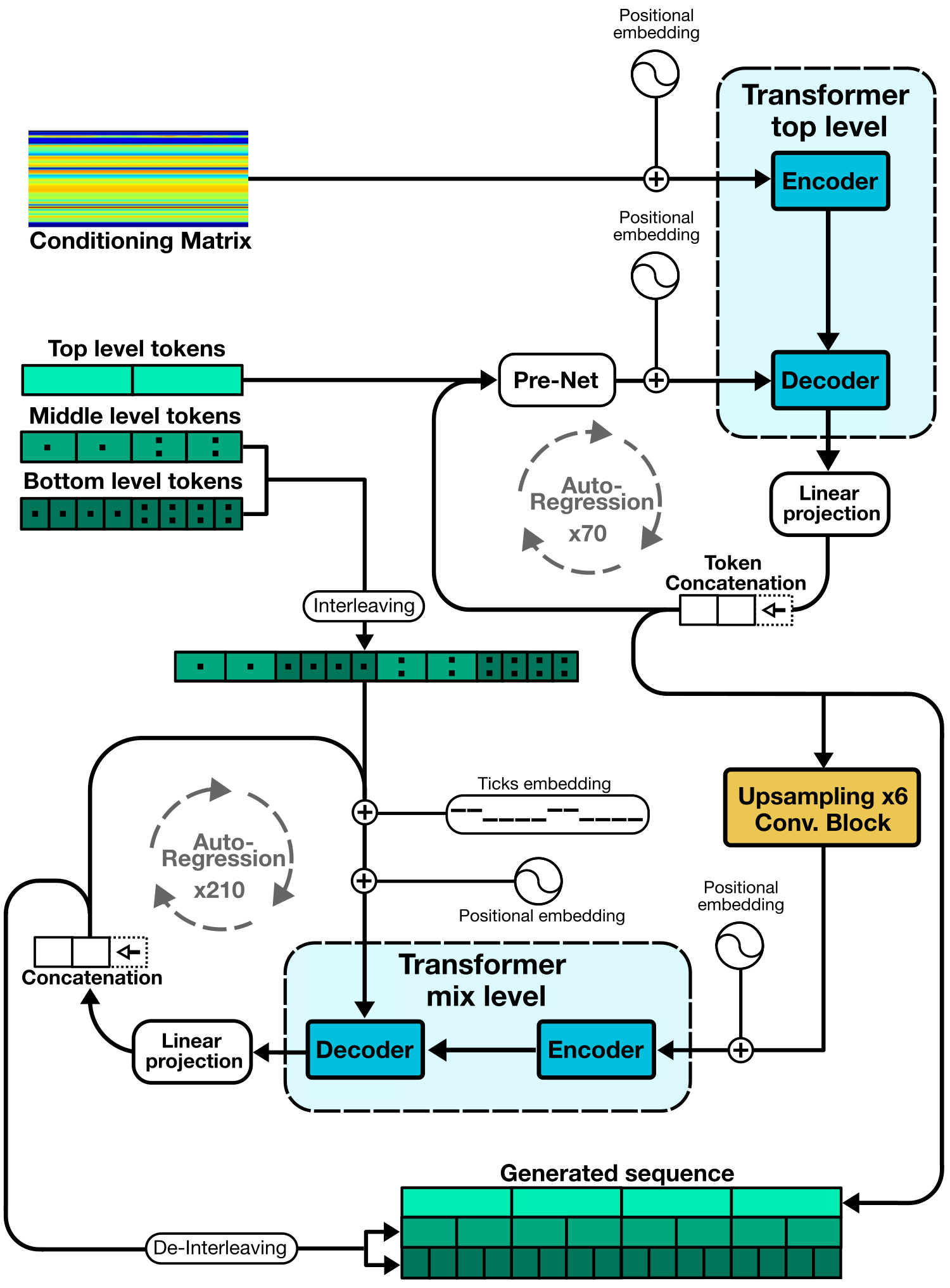}
\caption{The cascade of transformers for generating token sequences}
\label{fig:transformershema}
\end{figure}

As illustrated in Figure \ref{fig:transformershema}, the first transformer network receives the conditioning matrix as input to its encoder. The conditioning features extracted from this encoder are then passed as memory to the decoder. The decoder receives in parallel from the encoder memory a token primer. A linear projection layer at the output of the decoder predicts the next token using a softmax function. New tokens are iteratively re-injected into the decoder to produce a sequence of arbitrary length.
After generating a sequence of tokens, an upsampling block is used to scale the sequence, combining linear interpolation by a factor of 6 with convolution layers, to match the dimensions of the middle and bottom levels.

The second transformer's encoder takes the upscaled top tokens as input, and feeds this context to the decoder. The middle level and bottom level token seed are interleaved, before also being sent to the second transformer's decoder. This network handles the extension of the token sequence in the same way as the first decoder, alternating generation between middle and bottom levels by using tick embedding. 

We consider \ac{MER} signals to be a continuous, the endpoint of which depends on the recording process, not some feature of the signal itself. Consequently, we do not use a stop token, but an arbitrary limit of 3 seconds. When synthesizing new data, we evaluated the generation from either a start token, defined by the average of all the tokens in the dictionary or a primer taken from 3 seconds of pre-existing recordings.
Initial results led us to prefer the latter. We also considered that all the defining characteristics of a MER signal could be contained within a short signal window. So, to generate longer sequences, we designed a simple sampling method to limit decoherence in the token space. This method involves generating several token sequences in parallel from a shared primer, before appending them together into a long sequence. The junctions are then smoothed over by both the \ac{VQVAE} decoder and MelGAN which  generate smooth outputs. This parallelization also considerably reduces generation time. Finally, for all softmax layers, the nucleus sampling method was used with a cumulative probability of 0.9 during evaluation.
The networks were trained for 3000 epochs using an Adam optimizer, a learning rate of 1e-3, and a batch size of 64.

%% file: 04_M_M.tex
\section{Experiments} \label{sec:experiment}

\subsection{Qualitative validation - Human Perception}
\begin{figure}[t]
    \centering
    \includegraphics[width=8cm,keepaspectratio]{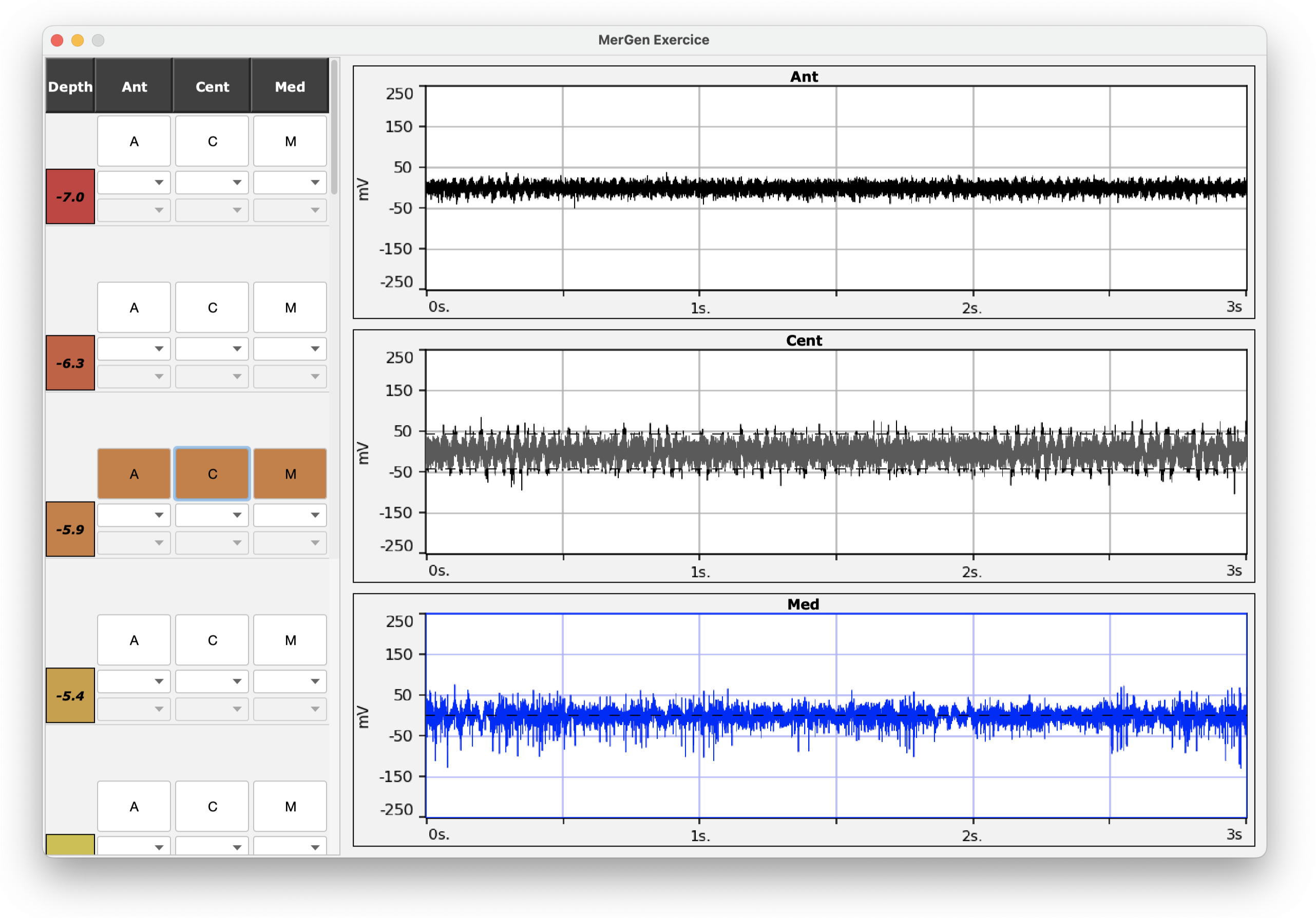}
    \vspace{-4mm}\caption{Human Perception Study Interface}
    \label{fig:interface}
\end{figure}

To validate the perceptual realism of the synthetic data, given the complexity of \ac{MER} signals, we created an interface (shown in Figure \ref{fig:interface}) similar to that used during \ac{DBS} interventions. This interfaces allows the user to listen to the signal and visualize its amplitude. 
The participants indicated:
\begin{itemize}
    \item whether they perceived the signal as being real or synthetic (both novices and experts); and
    \item the level of activity characteristic of the \ac{STN} using the annotation scheme from in Table \ref{table:annotation_prevalence} (experts only). 
\end{itemize}
The first indicates the signals' realism as perceived by both novice users and expert neurophysiologists. The latter's domain-knowledge and experience may allow them to be more capable of distinguishing real signals from generated ones but also allow them to be more distracted by meaningful variability within the dataset. The second allows us to determine if the generated signals reflect a particular context and to estimate how reproducible these annotations are on real data. By including the original expert annotator, we can measure both intra- and inter-operator variability. 
Novices were excluded due to a lack of necessary expertise.

In order to provide a reference for the participants, an entire real and simulated trajectory with original annotations is shown before the exercise. In the study, each participant was given three channels for two trajectories from the testing set, with a total of 131 recordings. A randomly-distributed half the signals were from the actual \ac{DBS} intervention with the other half generated using conditioning from the corresponding real signal. To improve the intra-operator estimate, the original annotator was given three additional trajectories.

\subsection{Qualitative validation - Data Augmentation}

Another validation would be to use MerGen for data augmentation and measure if it improves performance of other tasks. Realistic synthetic data should improve machine learning models by increasing the size and variability of the training set. In order to validate this, we reproduced our previous experiment on MER signal classification \cite{peralta2020sepaconvnet} on this new dataset, using one-second signals as input for binary classification. (The $++$ and $+++$ annotations are considered inside the \ac{STN} and the rest outside for the purposes of binarization.)
The following set-ups were used:
\begin{itemize}
\item  \ac{TRTR}; 
\item  \ac{TSTR}; and
\item  \ac{TRSTR}.
\end{itemize}
For the real data, a one-second window was randomly selected from the 3-second signal. For synthetic data, a 1-second window was randomly selected from the 10-second synthetic signals. The use of longer signals increases the likelihood of picking windows with degenerating coherence, a common problem for time-series generation. MerGen's training dataset was also used for the classifier. The testing set patients were used to evaluate the classifier's \ac{BACC}.

%% file: 05_Results.tex
\section{Results}

\subsection{Qualitative verification}\label{sec:qual_res}

\ac{MER} signals reflect the electrical activity of the surrounding anatomy, specifically: Action Potentials from one or more proximal individual neurons within a range of 200µm; and \ac{BUA} resulting from the activity of the surrounding tissue further from the electrode.
Acquisition artefacts, such as burst-like mechanical artifacts or 50-60Hz alternating current oscillations from tools in the surgical scene, are often also present. All these phenomena should be taken into account for realistic simulation, especially as they are discernible by ear and visible on a signal amplitude graph.

\begin{figure}[!t]
\centering
    \includegraphics[width=0.895\textwidth,keepaspectratio]{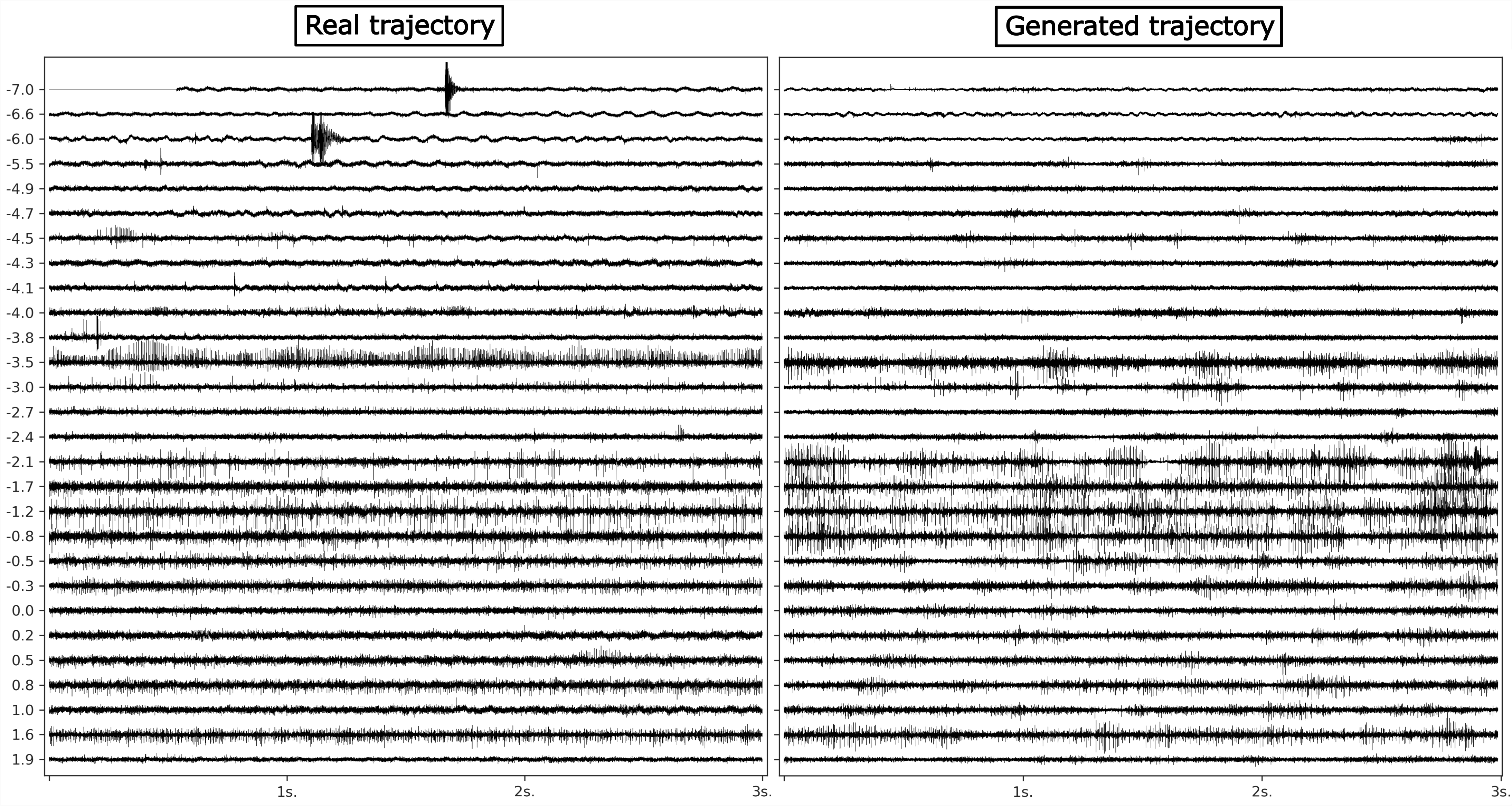}
    \caption{Real (left) and simulated (right) \ac{MER} signals from a single electrode trajectory. The conditioning information for the simulated data was extracted from the real signal to ensure they have the same context.}
\label{fig:realvsfake}
\end{figure}

To verify that MerGen reproduces these features, we visualised a set of 28 real MERs alongside corresponding simulated ones as shown in Figure \ref{fig:realvsfake}. These signals arise from different depths along the electrode trajectory which have varying conditioning matrices. The synthetic signals were generated using said matrices. The network generated a total of 1 minute and 24 seconds of simulated \ac{MER} signals in 11.3 seconds, using a RTX5000 graphics card.

The intensity of the simulated \ac{BUA} is often constant, which is not the case in real data. For the pre-STN white matter, the low-frequency artifact oscillation (10-20Hz) present in real signals is reproduced, but not the highly localized mechanical artefacts. Regarding neurons closer to the probe, the simulation mimics their abrupt activity observed in the real data. However, there is a limitation in terms of spike rhythm. \textit{In vivo}, spike activity often occurs in bursts following a pseudo-biological rhythm. This characteristic is not fully replicated in the simulated data, where this activity is more frequent and has a less structured rhythm.
Overall, there is significant but not perfect consistency between the real and simulated signals, making them largely coherent with the underlying anatomy.

\subsection{Qualitative validation - Human Perception}

Table \ref{table:human_res_1} shows the performance with which human users can distinguish between real and simulated signals. For novices, the accuracy, sensitivity, and specificity are clearly equivalent to chance. The experts ability to distinguish between real and generated signals is also close to chance, with accuracies between 35\% and 63\% depending on the expert. In addition, the low sensitivity suggests that experts are generally confident in the realism of the simulated signals.

\begin{table}[t]
\caption{Results for distinguishing between real and synthetic MER.}
\begin{tabularx}{\linewidth} { 
   >{\raggedright\arraybackslash}p{0.15\linewidth} 
  | >{\centering\arraybackslash}X 
  | >{\centering\arraybackslash}X 
  | >{\centering\arraybackslash}X }
\toprule
 & Novice & Expert & Original Annotator\\
 & $(n=11)$ & $(n=4)$ & $(n=1)$\\
\hline
Accuracy & 49.5\% $\pm$ 7.4\%  & 51.9$\pm$ 9.9\% & 54.1\% \\
Sensitivity & 46.9\% $\pm$ 8.7\%  &37.5$\pm$ 12.5\% & 45.2\% \\ 
Specificity & 52.0\% $\pm$ 8.9\% & 65.7$\pm$ 17.2\% & 63.1\%\\ 
\bottomrule
\multicolumn{4}{p{251pt}}{The Expert group includes all experts, including the original annotator.}
\end{tabularx}\vspace{0.5mm}

\label{table:human_res_1}
\vspace{-3mm}
\end{table}
 
\begin{figure}[t!]
\centering
    \includegraphics[width=.965\textwidth,keepaspectratio]{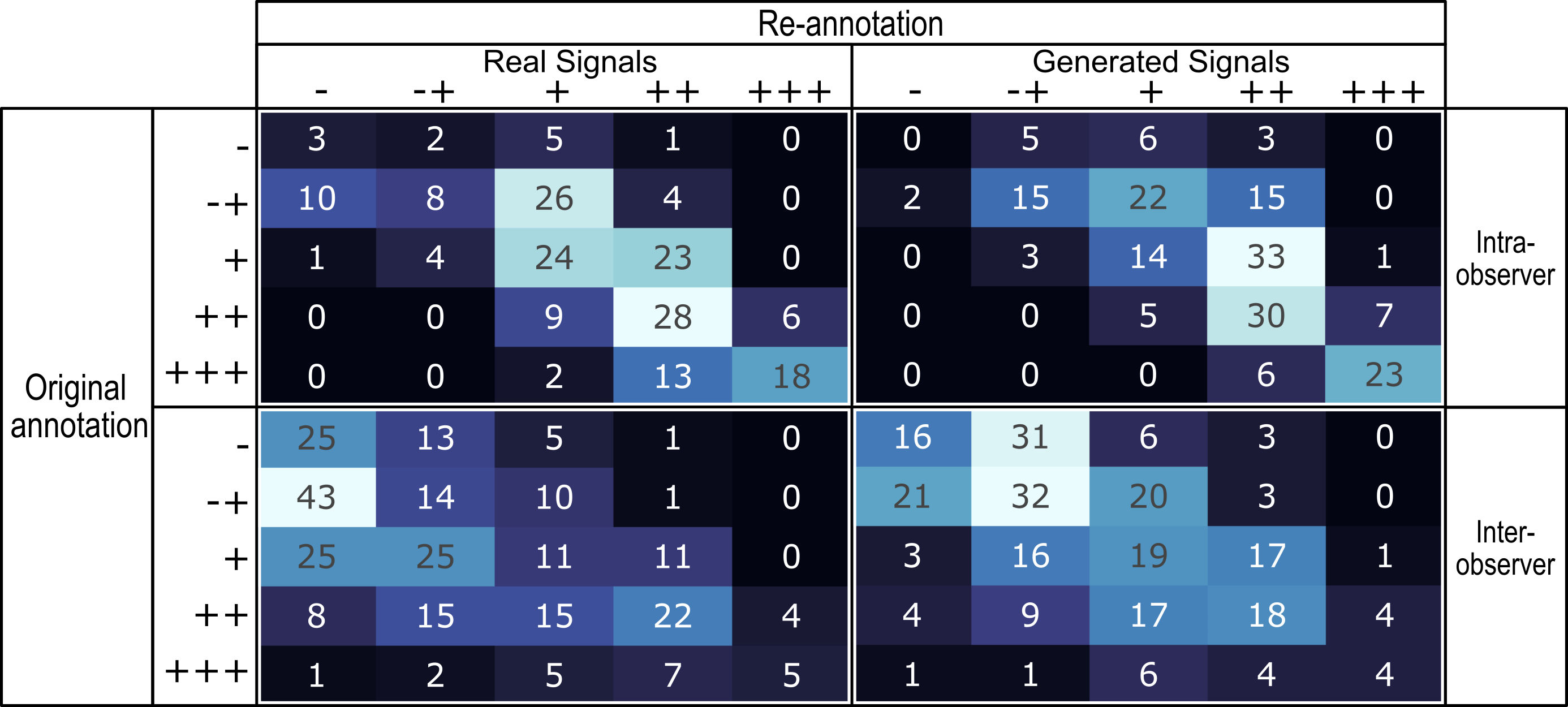}
    \vspace{-3mm}\caption{Confusion matrices showing the intra- and inter-observer variability for real and synthetic signals. 
    }
\label{fig:conf_mat_experts}
\end{figure}

The confusion matrices for estimating the annotation variability are shown in Figure \ref{fig:conf_mat_experts}. 
The intra-observer results for real signals were consistent with the originally annotated level of activity although with significant confusion between adjacent (and therefore similar) annotation values explained by the subjectivity of the task. These results were reproduced on the generated data, indicating that the conditioning induces signals that are coherent with a particular physiological context. With regards to inter-expert annotations, we note that the variability is slightly higher than the intra-observer variability, though still coherent. There is also a high degree of similarity in the annotations between synthetic and real data although, this is tempered by the slight increase in perceived activity in synthetic signals, leading to a rightward shift in their confusion matrices but slightly lower variability.
The intra-observer and inter-observer variability estimates suggest the generated signals are perceptually physiologically correct.

\subsection{Qualitative validation - Data Augmentation}

\begin{figure}[t!]
\centering
    \includegraphics[width=0.5\linewidth,keepaspectratio]{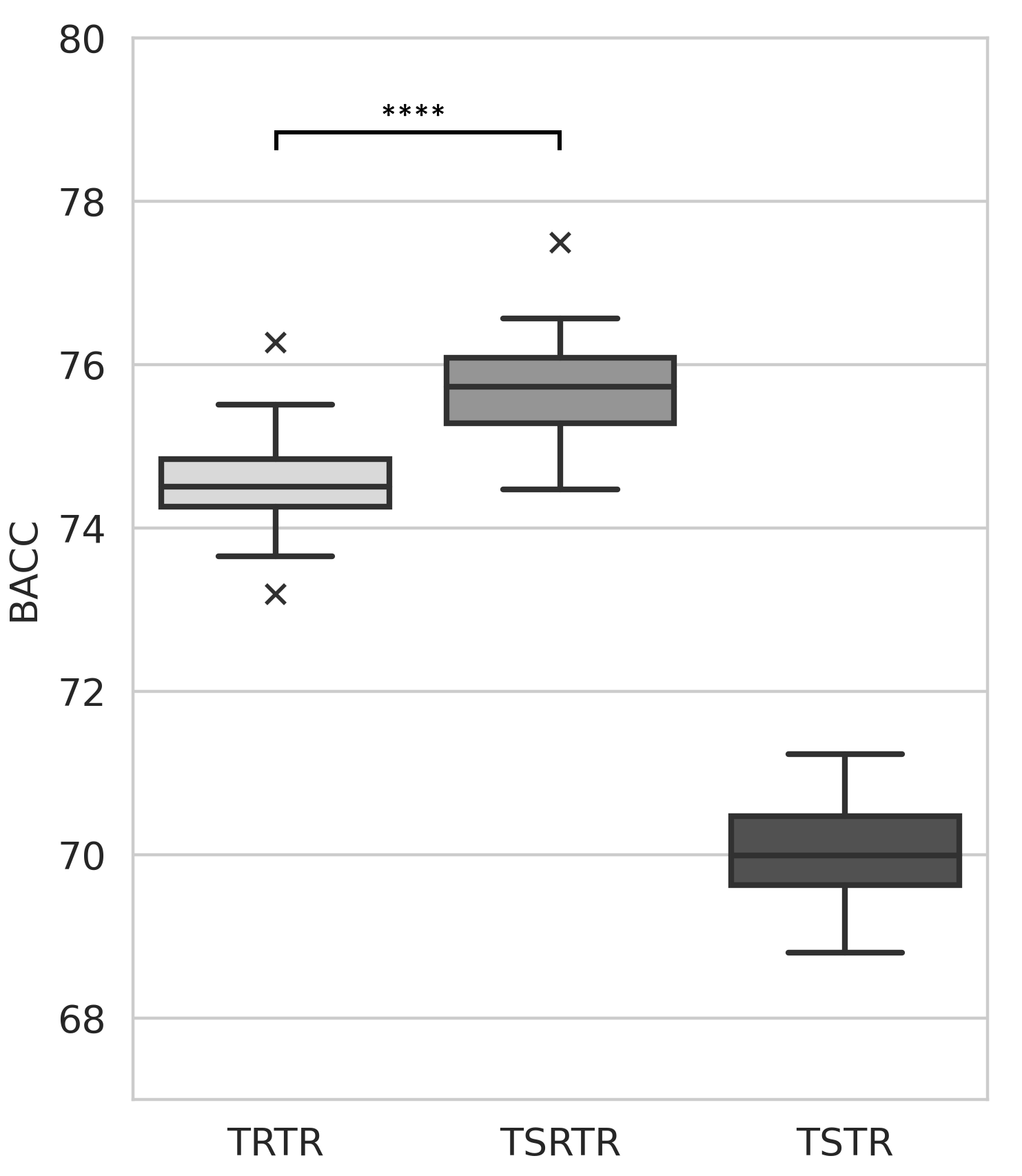}
\vspace{-3mm}\caption{Accuracy of a convolutional network for binary classification of signals according to their \ac{STN} activity level. Training is performed on real data (\ac{TRTR}), synthetic data (\ac{TSTR}), or both (\ac{TRSTR}). 
}
\label{fig:scn_data_augmentation}
\end{figure}

Figure \ref{fig:scn_data_augmentation} shows the results of the \ac{STN} classification experiment which are consistent with results regarding the reproducibility of \ac{MER} classification \cite{martin2024generalisation}.
Using both real and synthetic training samples (TSRTR) resulted in the highest performance, significantly above using only real data (TRTR). This indicates that the synthetic \ac{MER}s have a high degree of realism and variability although not enough to fully replace real data altogether. This confirms our qualitative observations regarding distribution differences between real and simulated \ac{MER} signals and presents some possibilities for improving \ac{DBS} electrode placement \cite{koirala2020mapping}.

%% file: 06_Discussion.tex
\section{Discussion}

\subsection{Analysis of the latent space}
\begin{figure}[t!]
\vspace{0mm}
\centering
    \includegraphics[width=.625\linewidth,keepaspectratio]{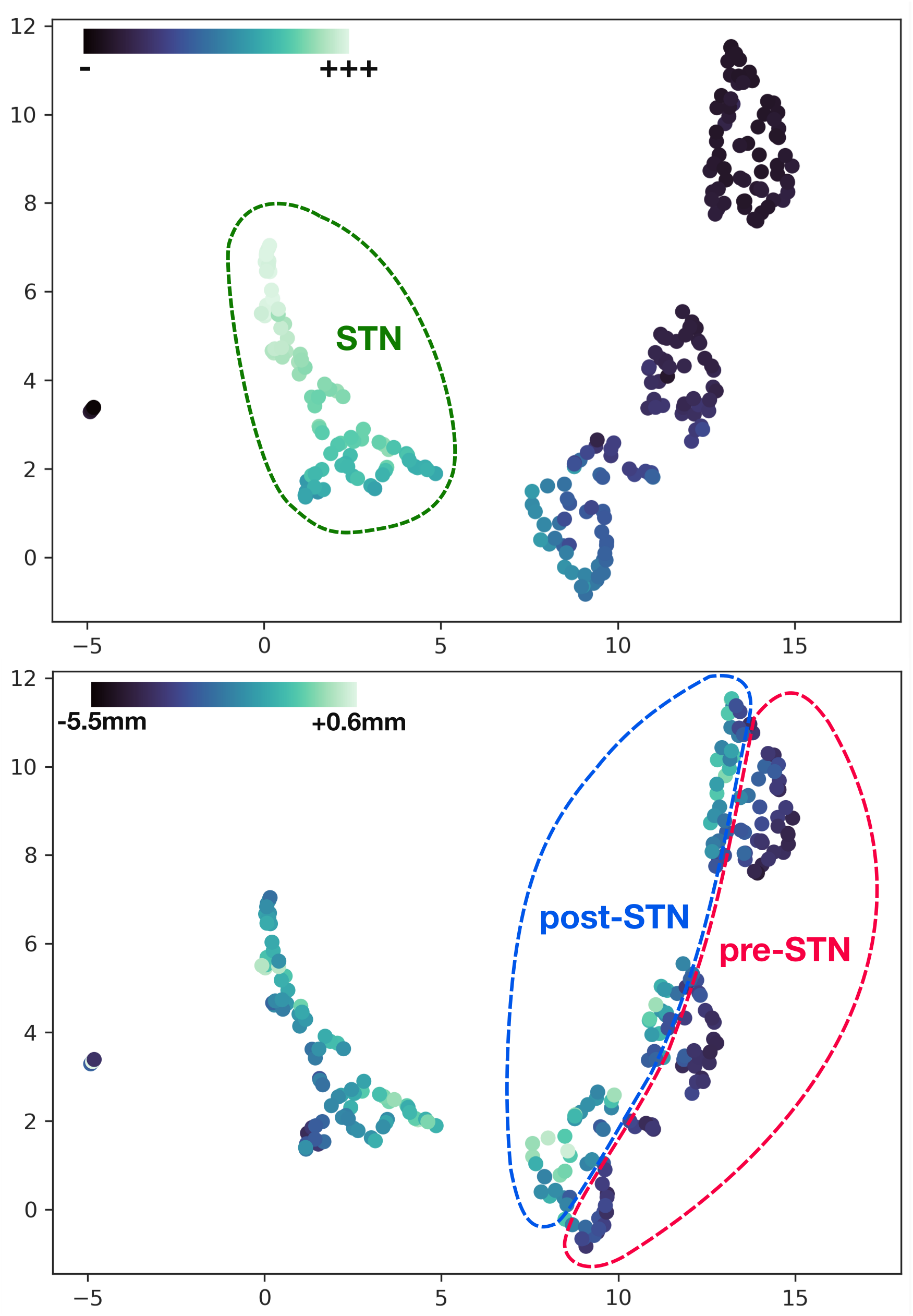}
\vspace{-3mm}\caption{UMAP dimensionality reduction of the \ac{VQVAE} top latent space embeddings. The top figure colours each point based on the average activity annotation, while the bottom figure uses the average estimated distance to target (EDT) associated with each token.}
\label{fig:umap}
\end{figure}

In order to visualise the structure of the latent space learnt by the \ac{VQVAE}, Figure \ref{fig:umap} shows the results of \ac{UMAP} (3 neighbours, minimal cosine distance of 0.1) performed on the tokens learnt by the topmost layer. The tokens form clear clusters which can be explained by the annotation being part of the conditioning information as shown in Figure \ref{fig:umap} (top). These annotations however do not distinguish between the substantia nigra (post-STN anatomy) and the pre-STN white matter bundles. Thus, it is interesting that the latent space does capture this information within the non-STN clusters  as shown in Fig. \ref{fig:umap} (bottom). This \ac{EDT}-dependence also occurs within the STN, which may be due to its different functional zones. Despite the obvious correlation between \ac{EDT} and annotation, MerGen treats them as separately useful for signal synthesis.


\subsection{Signal generation quality over time}
As mentioned in Section \ref{transformer_section}, we use a simple sampling method to prevent decoherence in longer signals. However, this method may lead to some phenomena that are not biologically accurate. For example, it is possible for a particular phenomenon (such as an artifact) to be periodically repeated producing patterns around $\nicefrac{1}{3}$ Hz (as each sub-sequence corresponds to 3 s of generated signal) which are not physiological and could confuse trainees.
In addition, longer patterns with frequency ranges lower than $\nicefrac{1}{3}$ Hz would be disrupted by this sampling strategy. These patterns are known to be present in wider-scale brain networks \cite{wu2008frequency}. Given that the signals acquired during \ac{DBS} electrode implantation are short, detecting said patterns is unlikely to impact a neurophysiologist's appreciation of the underlying anatomy, but lacking them could still limit realism.

\subsection{Limitations and future work}
The most immediate area of future work is the integration of MerGen into a \ac{DBS} electrode implantation training simulator. This is a non-trivial task as said interventions require a team of clinicians with different roles, such as neurosurgeons, neurophysiologists, anaesthesiologists, and nurses. Such a simulation should also take into account that \ac{MER} signals are not the only intraoperative data modality, but include electrode depth, fluoroscopic imaging, and patient responses to trial stimulation. Thus, ecologically valid simulations should be multi-person and only use \ac{MER} generation as a part of a larger whole.

Despite its perceived realism, using MerGen for medical resident training requires MER synthesis for a wide array of patients. Even with conditioning, our study protocol does not prevent the possibility of mode collapse, \textit{i.e.} that the generator has not modelled all of the patients in the training dataset. This could limit its full potential as trainees could still become subconsciously habituated to particular patient characteristics.

Another limitation of MerGen is the use of a single \ac{MER} acquisition set-up. This may limit its didactic use as trainees would be exposed to a single type of \ac{MER} which may not fully translate to other centres. In the future, information regarding the precise hardware could be encoded with the conditioning information in order to allow for the simulator to produce signals representative of other centres. However, this would require significant data acquisition on a broad array of devices which is difficult to obtain. It is also unclear if such models would outperform multiple single-centre ones and whether such device variability would significantly improve trainee performance. This finding also raises the question of the potential contribution of synthetic MER signal generation to medical resident training. In this context, the added pedagogical value should be evaluated against conventional training using pre-recorded signals. 
This single-centre focus can also significant affect the generalisation capabilities of deep learning models using MerGen for data augmentation \cite{martin2024generalisation}.

%% file: 07_Conclusion.tex
\section{Conclusion}
In this paper, we present MerGen, an electrophysiological signal generator which is the combination of three deep learning models: a MelGAN model to translate signals from the spectrogram domain, a multi-resolution \ac{VQVAE} model for representing these spectrograms as a sequence of discrete tokens, and a cascade of transformers for generating said sequences. In order to evaluate the perceptual realism of the generated signals, a human perception study was performed with both novices and expert neurophysiologists --- neither could differentiate between synthetic samples and actual acquisitions. A second study evaluating the MerGen's utility for data augmentation verified that the realism also extends to the more objective task of automated \ac{MER} classification. To the best of our knowledge, this work is both (1) the first application of these models to \ac{MER} signals, and (2) the first multi-faceted quantitative assessment of synthetic signal realism.

This work is the first step in the development of collaborative training simulators for performing \ac{DBS} electrode implantation interventions, specifically the component involving the training of medical residents interested in interventional neurophysiological analysis.

\section*{Acknowledgments}
We would like to thank Association France Parkinson for their financial support of Thibault Martin's doctoral research. Additionally, we are grateful to Jérôme Coste (Clermont-Ferrand University Hospital), Laurent Goetz (Adolphe de Rothschild Foundation Hospital), and Teodor Danaila (Lyon University Hospital) for their expertise.